\documentclass[11pt]{article}
\usepackage{soul}
\usepackage{amssymb}
\usepackage{url}
\usepackage{amsfonts}
\usepackage{amsmath}
\usepackage[top=1.2in, bottom=1.2in, left=1.2in, right=1.2in]{geometry}
\usepackage{graphicx,color,psfrag}
\usepackage[latin1]{inputenc}
\usepackage{multirow}
\usepackage{rotating}
\usepackage{natbib}
\usepackage{threeparttable}
\usepackage{booktabs}
\usepackage{subfig}
\usepackage{titlesec}
\usepackage{array}
\usepackage{tikz}
\usepackage{hyperref}
\usepackage{enumitem}
\usepackage{titling}
\usepackage{bbm}
\usepackage{titling}
\usepackage{rotating}
\usepackage{authblk}
\usepackage{soul}
\usepackage{float}
\usepackage{lineno}

\usepackage{pdflscape}
\newcommand*{\figuretitle}[1]{%
	{\centering
		\footnotesize{#1}          
		\par\medskip}          
}

\hypersetup{
	pdftitle={Favelas 4D},    
	pdfauthor={},     
	colorlinks=true,       
	linkcolor=blue,          
	citecolor=black,        
	filecolor=black,      
	urlcolor=black           
}
 
\newcolumntype{L}[1]{>{\raggedright\let\newline\\\arraybackslash\hspace{0pt}}m{#1}}
\newcolumntype{C}[1]{>{\centering\let\newline\\\arraybackslash\hspace{0pt}}m{#1}}
\newcolumntype{R}[1]{>{\raggedleft\let\newline\\\arraybackslash\hspace{0pt}}m{#1}}

\titleformat*{\section}{\centering\Large\sc}
\titleformat*{\subsection}{\large\bf}

\renewcommand{\baselinestretch}{1.3} 
\title{Favelas 4D: Scalable methods for morphology analysis of informal settlements using terrestrial laser scanning data}
\author[1]{Arianna Salazar Miranda\thanks{Correspondence: ariana@mit.edu}}

\author[1]{Guangyu Du}
\author[1]{Claire Gorman}
\author[1,2]{Fabio Duarte}
\author[1]{Washington Fajardo}
\author[1]{Carlo Ratti}
\affil[1]{Senseable City Lab, Department of Urban Studies and Planning, Massachusetts Institute of Technology}
\affil[2]{Pont\'{i}ficia Universidade Cat\'{o}lica do Paran\'{a}, Curitiba, Brazil}

\begin{document}
\maketitle

\begin{abstract}
{\footnotesize One billion people live in informal settlements worldwide. The complex and multilayered spaces that characterize this unplanned form of urbanization pose a challenge to traditional approaches to mapping and morphological analysis. This study proposes a methodology to study the morphological properties of informal settlements based on terrestrial LiDAR (Light Detection and Ranging) data collected in Rocinha, the largest favela in Rio de Janeiro, Brazil. Our analysis operates at two resolutions, including a \emph{global} analysis focused on comparing different streets of the favela to one another, and a \emph{local} analysis unpacking the variation of morphological metrics within streets. We show that our methodology reveals meaningful differences and commonalities both in terms of the global morphological characteristics across streets and their local distributions. Finally, we create morphological maps at high spatial resolution from LiDAR data, which can inform urban planning assessments of concerns related to crowding, structural safety, air quality, and accessibility in the favela. The methods for this study are automated and can be easily scaled to analyze entire informal settlements, leveraging the increasing availability of inexpensive LiDAR scanners on portable devices such as cellphones. 
}

\end{abstract}
\thispagestyle{empty}
\setcounter{page}{0}
\clearpage

\section{Introduction}
Of the roughly four billion people who live in cities worldwide,\footnote{https://data.worldbank.org/indicator/SP.URB.TOTL} nearly one billion live in informal settlements as of 2020.\footnote{https://www.worldbank.org/en/topic/urbandevelopment/overview.} Informal settlements are aggregations of homes and businesses constructed by their residents, in an initially unplanned form of urbanization. Their complex morphology is the outcome of spontaneous and competitive building without official land tenure, which results in dense and multilayered structures built up around labyrinthine street networks. These environments tend to be insalubrious and unsafe compared to the formally developed parts of the city: their exclusion from urban infrastructure and official management exposes them to a variety of public health and security challenges, and their typical location on undesirable land can result in several forms of environmental precarity. As the growing urban population continues to outpace the construction of new affordable housing by official channels, these settlements are expanding around the world.\footnote{https://unstats.un.org/sdgs/report/2019/goal-11/}

Morphology is a productive angle to study informal settlements because it allows for a systematic understanding of their development patterns, character, and how they cope with change \citep{conzen_thinking_2004, moudon_built_1989, siksna_effects_1997}. For informal settlements in particular, measuring morphology is an entry point to broader inquiry about the tendencies of unfettered urban development and the challenges that attend it, including lack of accessibility and cadastral mapping, crowding, environmental health, and safety. Moreover, characterizing the morphology of informal settlements can inform effective urban design interventions and policies to alleviate the negative impacts of living in informal settlements. 

The irregular and complex morphology typical of informal settlements can act as a barrier to traditional forms of mapping and morphological analysis. Remote sensing data is available and useful for differentiating informal settlements from other land cover \citep{prabhu2021a, schmitt2018a}, but its resolution is insufficient for detailed study within individual settlements. Aerial data collection with photography or radar, though more granular than remote sensing data \citep{gevaert2017a, sliuzas2017a}, tends to suffer from occlusions and fails to capture the vast heterogeneity in materials, building types, and design that are typical of informal settlements. Finally, ethnographic or observational data gathered inside the settlements benefits from the richness and nuance of community engagement but is too labor intensive to yield more than a partial or piece-wise analysis in most cases \citep{cavalcanti2017a, mukeku2018a}. These challenges have left informal settlements understudied in proportion to their relevance to urban planning research.

This paper addresses these shortcomings by introducing a scalable method of quantitative morphological analysis for informal settlements using data obtained from three-dimensional LiDAR (Light Detection and Ranging) laser scanning. LiDAR scans provide data on the environment in the form of unordered point clouds. The core of our methodology is a procedure to extract geometric properties of streets and facades from unordered point clouds to analyze their morphology. The first part of the paper focuses on measuring the morphology of``street scenes,'' a spatial unit consisting of a street segment and all its facades. We propose five spatial metrics, which we show how to compute using the LiDAR data: \emph{street width; street elevation; facade heterogeneity}, a measure of inconsistency in facade height; \emph{facade density}, a measure of crowding among buildings; and \emph{street canyon}, a metric that captures the width and depth of the facade relative to the street. We use these metrics to furnish a \emph{global} analysis comparing different street scenes to one another. In the second part of the paper, we extend our analysis to a higher resolution by constructing \emph{local} measurements of the same metrics, at half-meter increments along the primary axis of each street scene, allowing us to trace the morphological variation within each street scene. Finally, we map our local results directly into plans of the favela, showing the nuances of morphological variation in context. These complementary resolutions provide detailed insights on the morphology of informal settlements, which could inform policy interventions designed to improve infrastructure or reduce environmental risk. Furthermore, the perspective of our terrestrial LiDAR data privileges the view from the street, enabling our analysis to capture the built environment from an embedded vantage point.

We implement our methodology using LiDAR data collected at the street level in Rocinha, the largest favela (informal settlement) in Rio de Janeiro, Brazil. The point cloud data we use consists of two sample LiDAR scans  collected in the year 2020 from separate areas of Rocinha, which capture a range of spatial conditions. The first scan corresponds to a formalized public plaza, which we refer to as the ``Courtyard Scan'' and the second scan corresponds to an informal residential area that we refer to as the ``Hillside Scan.'' We use the morphological variety across these two scans to demonstrate the efficacy of our morphological metrics for analyzing the spectrum of development forms present in the favela. 

Our analysis demonstrates how terrestrial LiDAR can be used to create scalable quantitative measures of morphology to study the spatial organization of informal settlements. In particular, we show that our measurements constructed using LiDAR data reveal meaningful differences and commonalities both in terms of the global morphology of the favela and in terms of the local variations within street scenes. As an expansive and consolidated favela, Rocinha makes a favorable object of study because it is sufficiently well mapped to offer ground truth for our analysis.\footnote{The community of Rocinha first emerged in 1927 and by 1933 comprised a group of 67 shacks. Today it is recognized as an administrative region of Rio de Janeiro (as of 1993) and may house as many as 150,000 people \citep{Fabricius2008}.} To validate that our measurements are indeed capturing meaningful morphological characteristics, we evaluate them according to the ground truth established by Google Earth and local maps created by the Public Works Company of the State of Rio de Janeiro (Empresa de Obras Publicas do Estado do Rio de Janeiro).

This paper complements current methods of quantitative morphological analysis for informal settlements using remote sensing data. Remote sensing-based studies have refined methods of identifying informal settlements and modeling their growth. \citet{wurm2019a} demonstrate the utility of satellite imagery for slum identification, showing that favelas in Rio de Janeiro delineated using visual assessment of satellite images matched the official boundaries listed in the Brazilian Census with an accuracy of almost 94\%. Complementary work uses machine learning models trained on satellite imagery to predict the presence and expansion of slums \citep{duque2017a, wurm2019a, stark2020a}. Morphologically, these studies focus on identifying the overall shape and size of informal settlement footprints by differentiating their texture from those of other types of land cover as it appears in 2D satellite image data. Our study extends this literature by differentiating between the distinct urban patterns within informal settlements, quantitatively analyzing the internal variations in urban texture across the favela beyond its overall morphological differences from other forms of development.

A second related category of research focuses on the definition and classification of informal settlement topologies using street network maps. This literature aims to analyze the morphology of informal settlements either in contrast to that of the formal city or as a globally consistent urban typology of its own \citep{sobreira2007a}. Analytical methods that produce wide-ranging taxonomies of urban plans \citep{barthelemy2014a, boeing2021a, taubenboeck2018a} are replicated in studies that focus on classifying the street networks of informal settlements \citep{zappulla2014a, loureiro2017a} or comparing them to other topological conditions such as those of the medieval Portuguese towns that could be considered their ancestors \citep{loureiro2019a}. Work on this topic addresses morphological trends and characteristics within informal settlements, considering more detail and internal variation than remote sensing-based studies do, while still reducing the built environment to its plan. In slums and informal settlements in general, but especially in favelas, vertical expansion is a key architectural dynamic: families build new floors on top of existing structures as relatives move in, or occasionally sell roof space to another tenant to build upon as if it were empty land. Topological studies that only address street networks deemphasize this key dynamic of informal settlement construction. Our analysis addresses factors related to topology, such as street width, while approaching the built environment with attention to 3D factors that topological analysis typically does not consider, such facade height and street canyon.

This study also contributes to a third category of research that focuses on studying informal construction at the building scale. This literature has focused mostly on approaches derived from concepts like pattern language \citep{alexander1977a} and shape grammar \citep{stiny2015a}, which have been applied to analyze patterns of construction in informal settlements \citep{kamalipour2020a, dovey2020a}, occasionally for the purposes of growth modeling or parametric design \citep{dovey2013a, bardhan2018a, chokyu2018a, verniz2020a}. While these lines of inquiry address 3D morphology in great detail as it relates to the establishment and structural evolution of individual buildings, they are difficult to scale to entire settlements. Our study similarly focuses on 3D morphology and extends this approach by considering building-scale attributes and contextualizing them within their larger volumetric and topographical environment through the combination of the local and global analysis.

Finally, our work also contributes to a limited selection of LiDAR-based analyses that have been conducted in favelas at small scales. These approaches rely on manual data processing, making them unsuitable for analyzing large areas. Research in this area has primarily focused on delineating roof boundaries for Digital Elevation Model (DEM) development and counting the number of units in buildings for cadastral surveying \citep{temba2015a, ribeiro2019a}. This paper presents a method to analyze the morphology of informal settlements at scale, which allows us to move beyond case study evidence when sufficient LiDAR data is available. 

The paper is organized as follows. Section \ref{data} discusses the LiDAR data. Section \ref{metrics} describes how the morphological measurements taken in each LiDAR sample are calculated. Section \ref{analysis} details the global and the local morphometric analysis, summarizes the results in morphological maps of the favela, and validates the measurements they are based on. Section \ref{discussion} discusses the significance of these results in relation to the real urban conditions of Rocinha. Finally, in section \ref{conclusion} we discuss the conclusions that can be drawn from our analysis, their contribution to the goal of understanding informal settlements worldwide, and the opportunities they open for future study.

\section{Data Description}\label{data}
To analyze the morphology of street scenes we use data obtained from LiDAR: a 3D scanning technology that captures laser range measurements by measuring the level to which a laser pulse emitted from the tripod-mounted scanner reflects its beam. The resulting data are a set of individual distances stored as points with 3D coordinates registered spatially using the GPS position of the sensor, also known as a point cloud.\footnote{Each point also stores information including its normal (the vector perpendicular to it), an intensity value that measures the strength of the reflected laser beam, and an RGB color derived from a separate process of photogrammetry.} Our data was gathered by a company that offers laser scanning services in Brazil.\footnote{For more details on the company, see https://www.brtech3d.com.br/ [ scanned at a rate of 750,000 points per second, from a total of 22 survey points]}

Unlike many point clouds used for urban research, which rely on aerial LiDAR scanning for tasks like building delineation \citep{temba2015a}, the LiDAR data used in this study was taken by a team of technicians working terrestrially to capture the streets of Rocinha. The main advantage of using terrestrial over aerial LiDAR data in this context is coverage. Given the steep and narrow roads in the favela, it is challenging to collect clear and comprehensive data using mobile LiDAR scanning via transportation modes such as cars or buses. For instance, our scanned areas include a park and a narrow exterior staircase, two areas that only foot traffic could penetrate. Access to pedestrian-only areas of the favela is fundamental to our ultimate objective of comprehensively understanding its built environment: currently, only 23\% of Rocinha's total road distance is included in Google Street View, because only that fraction of the settlement could be smoothly captured from a car. Terrestrial LiDAR scanning allows us to measure the morphology of areas beyond the small and disproportionately formalized portion of Rocinha previously available for study. An important consequence of the terrestrial scanning approach on our data is that the point cloud includes relatively clear information for the ground and building facades, but it does not capture roof data for buildings reliably. For this reason, we focus our analysis on streets and facades rather than building footprints.

\begin{figure}[!ht]
	\centering
	\figuretitle{}
	\includegraphics[width=0.8\linewidth]{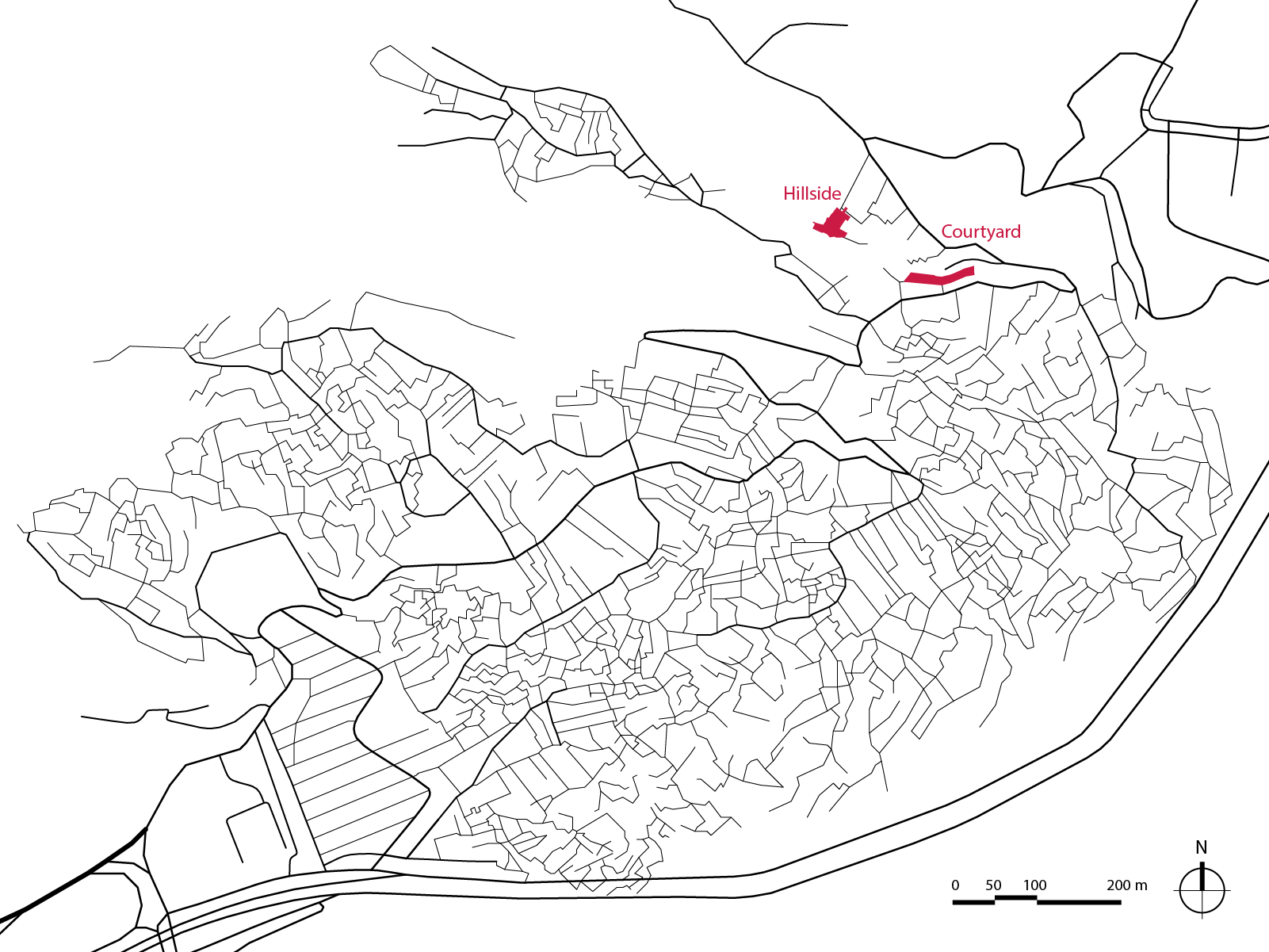}
	\caption{Map showing the scan locations.}
	\label{scan_locations}
\end{figure}

The scanning sample for this project was collected in two separate parts of Rocinha: a Courtyard and a Hillside that cover about 480 square meters of ground in total. Figure \ref{scan_locations} shows the locations of both scans. The Courtyard Scan is located below a main road, and the Hillside Scan is near it, further up the hillside. Both scans are on the upper and outer fringes of Rocinha, representing the range of conditions possible within Rocinha's informal development. The urban environment of the Courtyard is more formalized than is typical of Rocinha, but it is useful as an analytical starting point and a comparison to the Hillside scan. It represents a public plaza called Praca da Vila Cruzada, formalized and renovated to be the site of a tourist center and handicraft fair due to its panoramic views of the favela.  In contrast to the Courtyard Scan, the Hillside Scan captures the compact, layered morphology typical of Rocinha's piecewise development: it follows a staircase that strings together a conglomeration of small buildings as it runs up a steep incline and intersects with a flat cross-street. This scan incorporates a range of built conditions common to Rocinha, including the steep public street, small facades, and covered areas. In this scan, the boundaries between public and private space are unclear. While the Courtyard is a distinctly public space, the Hillside Scan includes street scenes that represent residential buildings that open onto the enclosed staircase connecting them together. 

Prior to processing, the point cloud included a total of 116,772,368 points (Courtyard: 45,342,425 points, Hillside: 71,429,943 points), scanned from 22 different survey locations (Courtyard: 5 survey points, Hillside: 17 survey points). For the purposes of this study, we downsample the data to 50 mm resolution, reducing the point cloud to 2,071,769 points in total (Courtyard: 607,219 points, 99.15\% reduction; Hillside: 1,464,550 points, 96.8\% reduction). This lower resolution allows us to retain all the information relevant to the scale of the street while limiting the size of our data to the minimum required for the analysis.

To summarize the morphological properties of the favela, we divide it into a series of spatial units that we call ``street scenes.'' Each street scene includes a street segment and the building facades that directly face it. To extract street scenes from LiDAR point cloud data, we follow two main steps: (1) extracting the street segments and building facades (horizontal and vertical planes, respectively) from the point cloud, and (2) grouping street segments and their adjacent building facades together. In the subsections that follow, we explain these steps in more detail.  

\subsection{Extracting Planes from the Point Cloud}

\begin{figure}[!ht]
	\centering
	\figuretitle{}
	\includegraphics[width=0.8\linewidth]{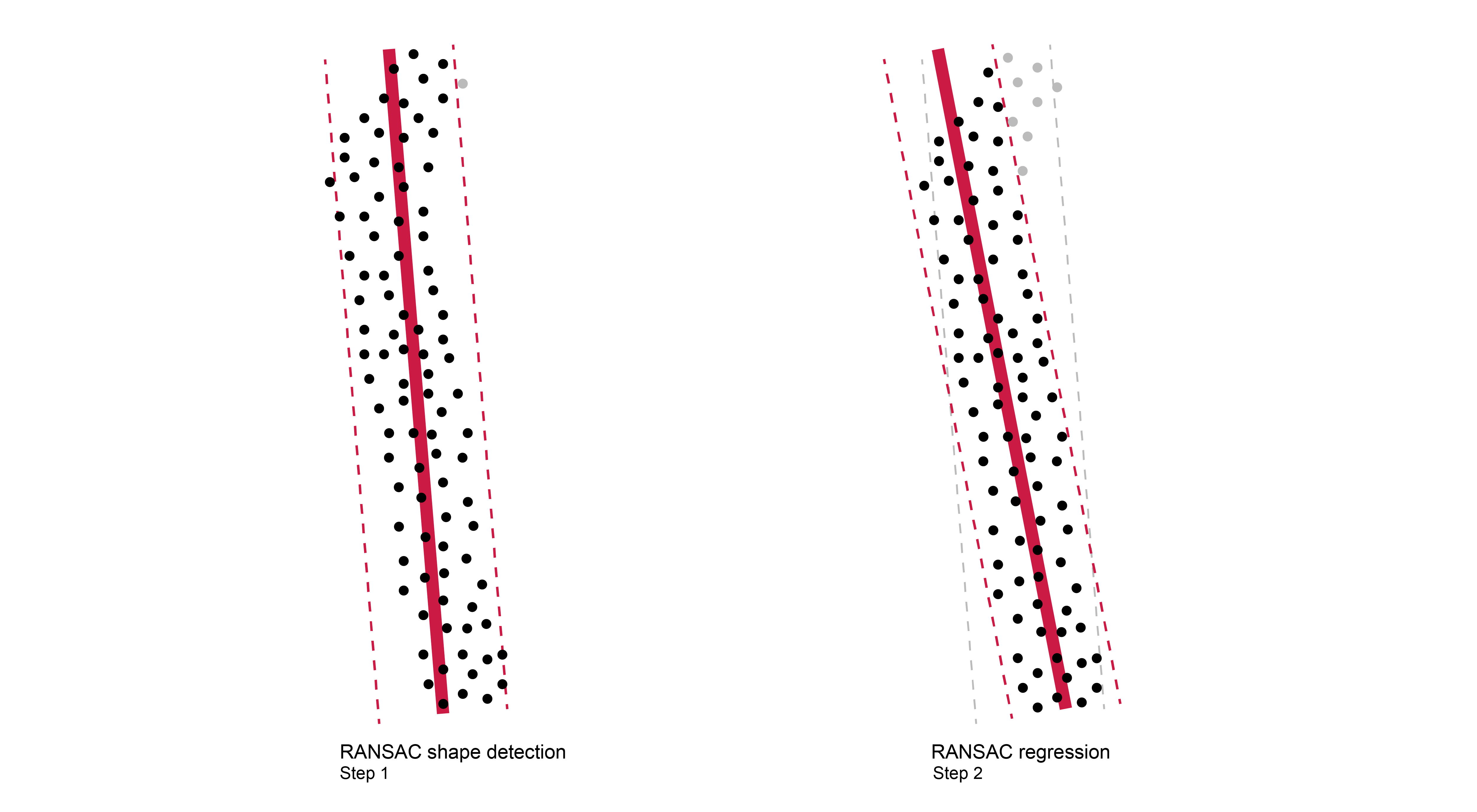}
	\caption{Graphic explanation of the plane extraction from point cloud shown as a section view of vertical plane. Left panel shows the RANSAC shape detection step. The right panel shows the RANSAC plane regression step.}
	\label{ransac}
\end{figure}

To extract horizontal and vertical planes from the LiDAR point cloud we use a two-part process of shape detection and plane regression. Both steps use versions of the Random Sample Consensus (RANSAC) algorithm, which is well suited to fitting models when there is a high proportion of outliers in the data \citep{fischler1981a}. In our case, these outliers come from trees, humans, and urban furniture that need to be filtered out of the point cloud before its built environment can be analyzed. Figure \ref{ransac} shows a graphical representation of the two RANSAC steps, with shape detection (step 1) on the left, and plane regression (step 2) on the right.

We first use the RANSAC shape detection method introduced by \citet{schnabel2007a} to extract shape models from unorganized point clouds.\footnote{\citet{schnabel2007a} adaptation of the RANSAC algorithm for shape detection uses a recursive process to find the best plane model. We set the distance $\epsilon$ to 0.005, and we allow the maximum deviation angle $\alpha$ to range between 12.5 and 35, to allow more variance on the size of the output planes. We set the minimum support points $\tau$ to 750 to include small planes from buildings but exclude other objects, such as urban furniture. See appendix for a detailed explanation of the parameters used.} The algorithm takes a point cloud as input (as well as a set of specific parameters\textemdash see Table \ref{ransac_table1} in the appendix) and outputs fragments of the point cloud that roughly fit the shape of a plane within a specified margin (Figure \ref{ransac}, left panel). Using this algorithm, we obtained 111 point cloud fragments in total: 52 fragments for the Courtyard Scan and 59 fragments for the Hillside Scan. These fragments vary widely in size, comprising anywhere from 807 to 136,599 points.

The second part of our plane extraction process uses RANSAC regression to establish a geometric plane model for each point cloud fragment detected in the first step. The objective of this step is to find the plane model that best fits each point cloud fragment (Figure \ref{ransac}, right panel), which is important because the coefficients of that plane model enable us to project the 3D point cloud fragment into 2D. Once we have projected the 3D points from a point cloud fragment into 2D using its regression plane, we can find the convex hull\textemdash a bounding shape\textemdash for the plane, which allows us to measure the geometric properties of the facade or street segment it represents.

\subsection{Constructing Street Scenes from Planes}
To aggregate the extracted planes into street scenes, we group each horizontal plane (street) with the nearest vertical planes (facades) above it, using a nearest-neighbors approach.\footnote{We only consider facades located at the same or higher elevation as the street in order to avoid analyzing roofs, which are also horizontal planes but whose nearest facades would be below them.} This step relies on categorizing each plane as either horizontal or vertical using their normal direction.\footnote{If the dot product of the regression plane's normal vector with the z vector is greater than 0.95, the plane is horizontal. If it is less than 0.05, the plane is horizontal.} Our street scenes are intended to be semi-continuous and overlapping: the same facade may appear in multiple street scenes when separate public areas exist on either side of the same wall. Also, they vary substantially in size, since a horizontal plane of any scale may be considered a street as long as it forms the ground plane corresponding to the nearest facades. In total, we identified 94 planes in the point cloud and organized them into 10 street scenes: 2 from the Courtyard Scan and 8 from the Hillside Scan.\footnote{Planes having a point density of less than 150 points per meter are considered noise and are not considered in the analysis. We extract 111 planes in total, classify 17 of them as noise, and proceed with the remaining 94.} Each street scene includes between 4 and 11 planes.

\section{Morphology Metric Calculation}\label{metrics}
\begin{figure}[!ht]
	\centering
	\figuretitle{}
	\includegraphics[width=0.8\linewidth]{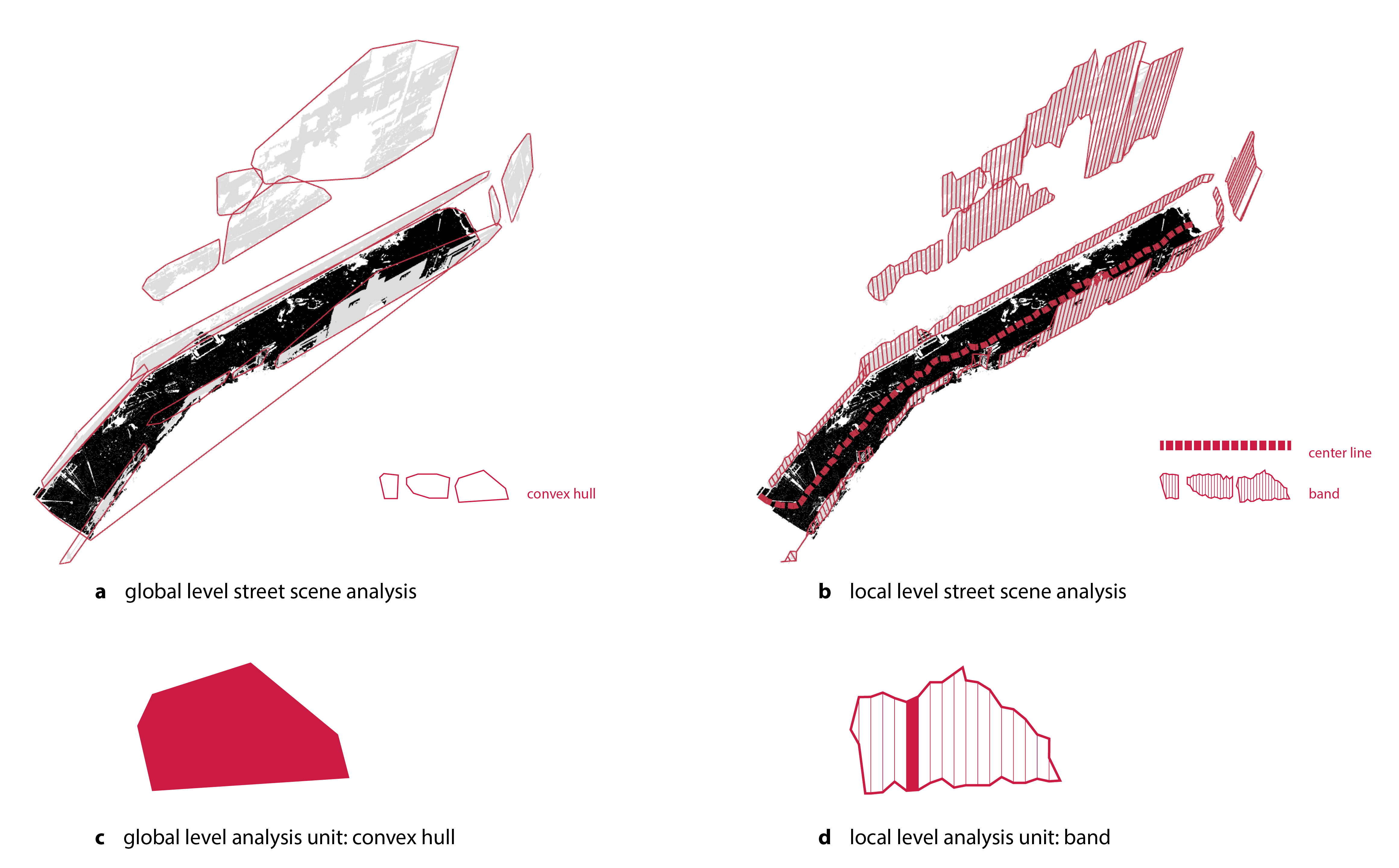}
	\caption{Figure shows a selected street scene depicting the components of the global and the local analysis. For global analysis, our calculations are based on the convex hulls of the planes in the street scene (left panel). For our local analysis, we divide each street scene into half-meter-wide bands (right panel). The two diagrams on the bottom of the figure represent the different resolutions achieved by this dual approach.}
	\label{global_local}
\end{figure}

This section describes our method for constructing morphological metrics for each street scene in our sample, at both global and local levels. The two scales of analysis are shown in Figure \ref{global_local}, which outlines the level of point cloud aggregation for each one. We compute a set of five metrics at each scale. These include street width and street elevation, which aim to describe the size and configuration of street scenes; and facade heterogeneity, facade density, and street canyon, which aim to describe the facade distribution and its relationship to the street. Our focus on these particular metrics draws on precedents in informal settlement studies that seek to identify and characterize unplanned areas using sets of quantitative measurements about the built environment \citep{kohli2012a, kuffer2014a}. We also draw from a subset of LiDAR literature that uses point clouds to extract key metrics related to urban morphology \citep{bonczak2019a, tooke2014a}. 

\subsection{Global Metric Calculation}
To calculate the global metrics, we take measurements for each street scene using the convex hulls of the planes that compose it (see Figure \ref{global_local}, left panel). In the following paragraphs we describe in detail how each metric is calculated. 

\emph{Street Width} is an important and applicable morphological metric, governing which and how many forms of transit can access a given location. We calculate it using a series of steps. To calculate overall street width for a street scene, we identify pairs of facades in the street scene that are effectively parallel (within 15 degrees) to one another. Then, we select the subset of these pairs that represent facades on opposite sides of the street from one another. To get the street width, we measure the perpendicular distance between these parallel facades that straddle the street.\footnote{ We cannot simply take the dimension of the street's convex hull in the X direction (``xdist'') because the inclusive nature of the convex hull inflates this value dramatically, especially when a street segment is curved.} If there are multiple pairs of facades that satisfy these calculations, we use the average of the distances between them. This metric relates to infrastructure and can be connected to connective bottlenecks in the favela that cause safety issues when essential and emergency services cannot access certain areas\citep{brelsford2018a}. 

\emph{Street Elevation} is intended to capture the overall topography of the point cloud and locate each street scene within it. It is calculated as the Z value of the centroid (middle point) of the horizontal plane identified as the street.\footnote{Based on the plane segmentation algorithm we use, this point exists in both the original point cloud fragment for the street and its estimated regression plane.} This metric has implications for safety and accessibility. Favelas tend to occupy steep, unstable land unsuitable for formal development, making them vulnerable to landslides and structural compromises associated with their unforgiving topography \citep{sandholz2018a}. The constant elevation changes in these settlements pose a major obstacle to the roughly 100,000 elderly people living in Rio's favelas as of the 2010 census, who have trouble navigating uneven sidewalks and staircases, and who will be a larger segment of Rio's population than children under 15 by 2030 \citep{guimar2016a}. Identifying sharp elevation changes between street scenes is a step towards efficiently improving connectivity and safety in the favela.

\emph{Facade Heterogeneity} in a street scene is a representation of the uniformity or disorder among building heights in the street. It is calculated using the standard deviation of vertical plane heights in each street scene. This metric is useful in various contexts. It can be applied to structural stability analysis, where risk can be correlated to building height \citep{shan2019a}; and it can also be connected to population density \citep{jedwab2021a} and microclimate in an area \cite{kakon2010a}. Moreover, this metric can also be related to the degree of informality in an area of the built environment: low facade variance indicates the standardized construction of large formal interventions like public housing projects, while high facade variance is associated with buildings in various stages of informal vertical expansion \citep{a2019a}. 

\emph{Facade Density} is calculated as the number of facades in a street scene divided by the surface area of the street they surround. Facade density is a proxy for crowding in the favela and is an important indicator in this analysis because it distinguishes between street scenes with a small number of large buildings from street scenes with many small ones.

Finally, we calculate \emph{Street Canyon}, which is the ratio of the maximum facade height in the scene to the width of the street. This spatial relationship has implications for the fresh air and sunlight that can reach certain areas, reflecting the level to which a given street scene is enclosed. This metric signals other issues at work in the built environment of a favela, since airflow and ventilation are highly related to urban morphology and have significant effects on pollution exposure \citep{jana2020a}. This issue is exacerbated in informal settlements, which tend to have higher levels of heat exposure than formalized urban areas \citep{scott2017a, wang-a}.

\subsection{Local Metric Calculation}
We now turn to discuss the adaptation of each calculation to collect the same measurements for the half-meter-wide bands of points that constitute our local analysis. This process requires slight modifications to some of the metric calculations. To calculate the local metrics at a higher resolution, we subdivide the point cloud into smaller segments. In particular, we subdivide the point cloud segments into half-meter-wide bands along the centerline (width) of each street scene, such that each band spans the street and includes segments of the facades on either side (see Figure \ref{global_local}, right panel). Finally, we perform the same two-step process of plane extraction (RANSAC shape detection and plane regression) for each band in order to delineate planes that represent the half-meter-wide street and facade segments within each band. We then calculate the morphological metrics for each band. 

Street width and street elevation are calculated exactly the same way for these bands as they are for entire street scenes. Instead of facade heterogeneity and facade density, which cannot be calculated at a scale smaller than a single facade, we construct a measure of facade height within each band by finding the height of the tallest facade segment on each side of the street and recording their average. If there are no facades on one side of the street, we simply record the facade height of the other side. We then use this facade height value to calculate our street canyon metric, the same way as for the global analysis. This results in a set of four metrics (street width, street elevation, facade height, and street canyon) that can be measured at half-meter intervals along every street scene to indicate morphological changes within it.

\section{Morphometric Analysis}\label{analysis}
In this section, we use the metrics computed at the global scale to compare different street scenes of the favela to one another, and the ones computed at the local scale to unpack the variation of morphometric metrics within a single street scene (at half-meter resolution). 

\subsection{Global Analysis}
\begin{figure}[!ht]
	\centering
	\figuretitle{}
	\includegraphics[width=1\linewidth]{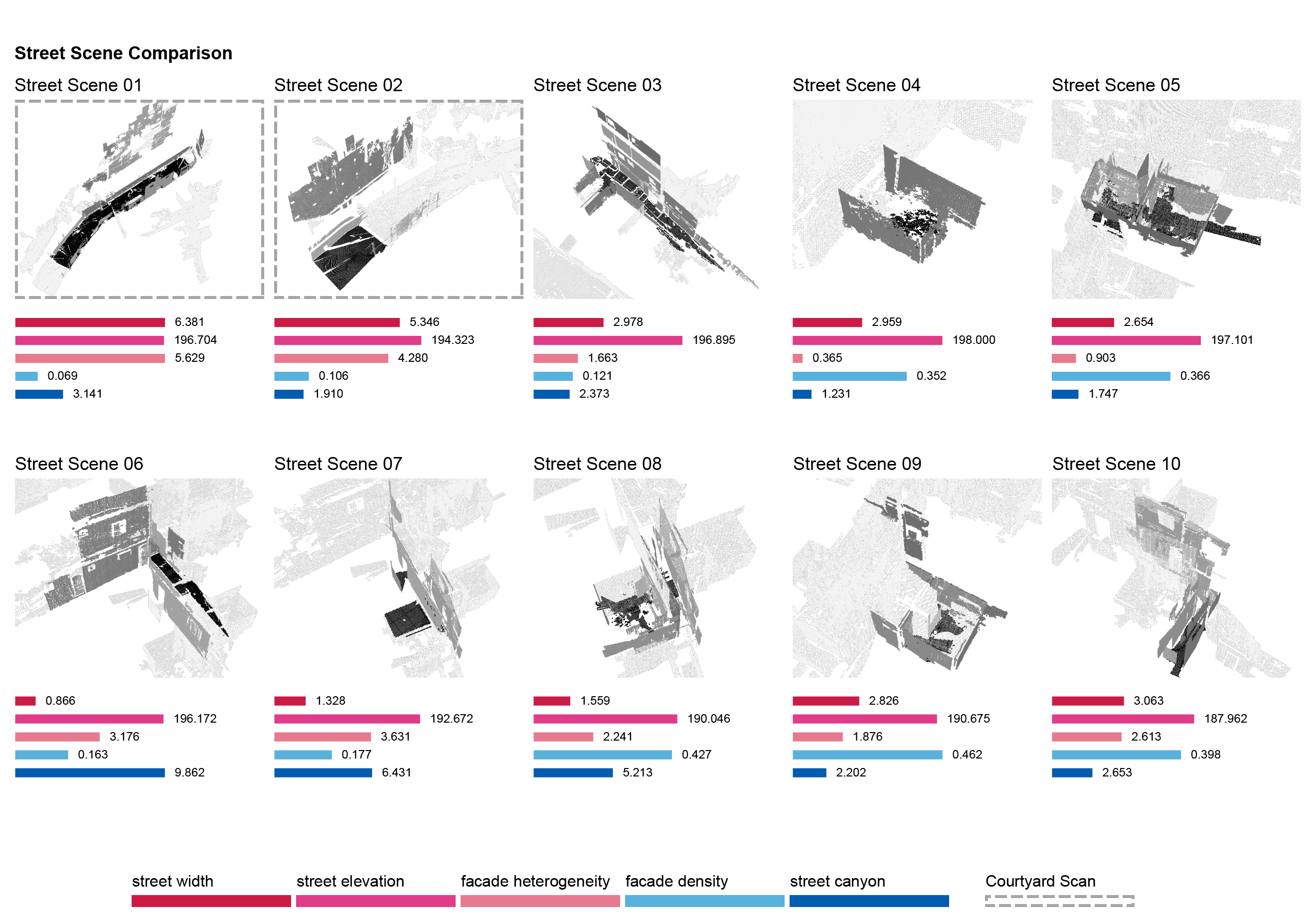}
	\caption{Figure shows the global analysis results for each street scene.}
	\label{global}
\end{figure}

Figure \ref{global} shows the 3D shape of each street scene and its values for the five morphological metrics (morphological profile) below it. The street scenes are numbered continuously, such that street scenes labeled with adjacent numbers are located next to each other in the favela. To ease the interpretation of the morphological metrics, we normalize each one relative to its maximum value. The figure shows the wide range of street types that can be found in the favela. For instance, more compact street scenes (street scenes 4, 6, and 9) tend to have the lowest street canyon values. In contrast, elongated street scenes characterized by tall facades and a narrow street (street scenes 3, 7, and 8) exhibit among the highest street canyon values. It is also apparent in this figure how different morphological dimensions vary together across street scenes: for instance, street scenes 1 and 2 have the highest values for street width, and facade homogeneity\textemdash so they might be considered thoroughfare street scenes. This pattern contrasts with the profiles of street scenes 8 and 9, which show some of the lowest values for those metrics and could be described as secluded, or low-traffic street scenes. A direct consideration of the similarities and differences among street scenes according to this global analysis can be achieved by visualizing each street scene as a point in 5 dimensions, where each dimension is a morphological metric (see Figure \ref{comparative} in the appendix). 

Beyond the typological similarities captured by the 5 morphology metrics, street scenes can also be analyzed in terms of their spatial similarities. As we will show, there are also strong similarities across neighboring street scenes that distinguish proximate streets from those located farther away. Street scenes 1 and 2 are located in the Courtyard Scan, so they share qualities of scale and spaciousness associated with the large, formalized public plaza they represent. Meanwhile, street scenes 3 through 10 are located in order of descending elevation in the Hillside Scan, and they exhibit the compact and variable nature of their more informal environment. The fact that our morphology metrics capture these differences is reassuring as it enables us to trace morphological trends in the favela that are spatially continuous as well as those that are not.

\subsection{Local Analysis}
Our global analysis shows how morphological metrics vary across the entire scanning area, indicating trends and relationships across all of the street scenes. In our local analysis, we go one step further to show how each morphological metric varies within individual street scenes.

\begin{figure}[H]
	\centering
	\figuretitle{}
	\includegraphics[width=1\linewidth]{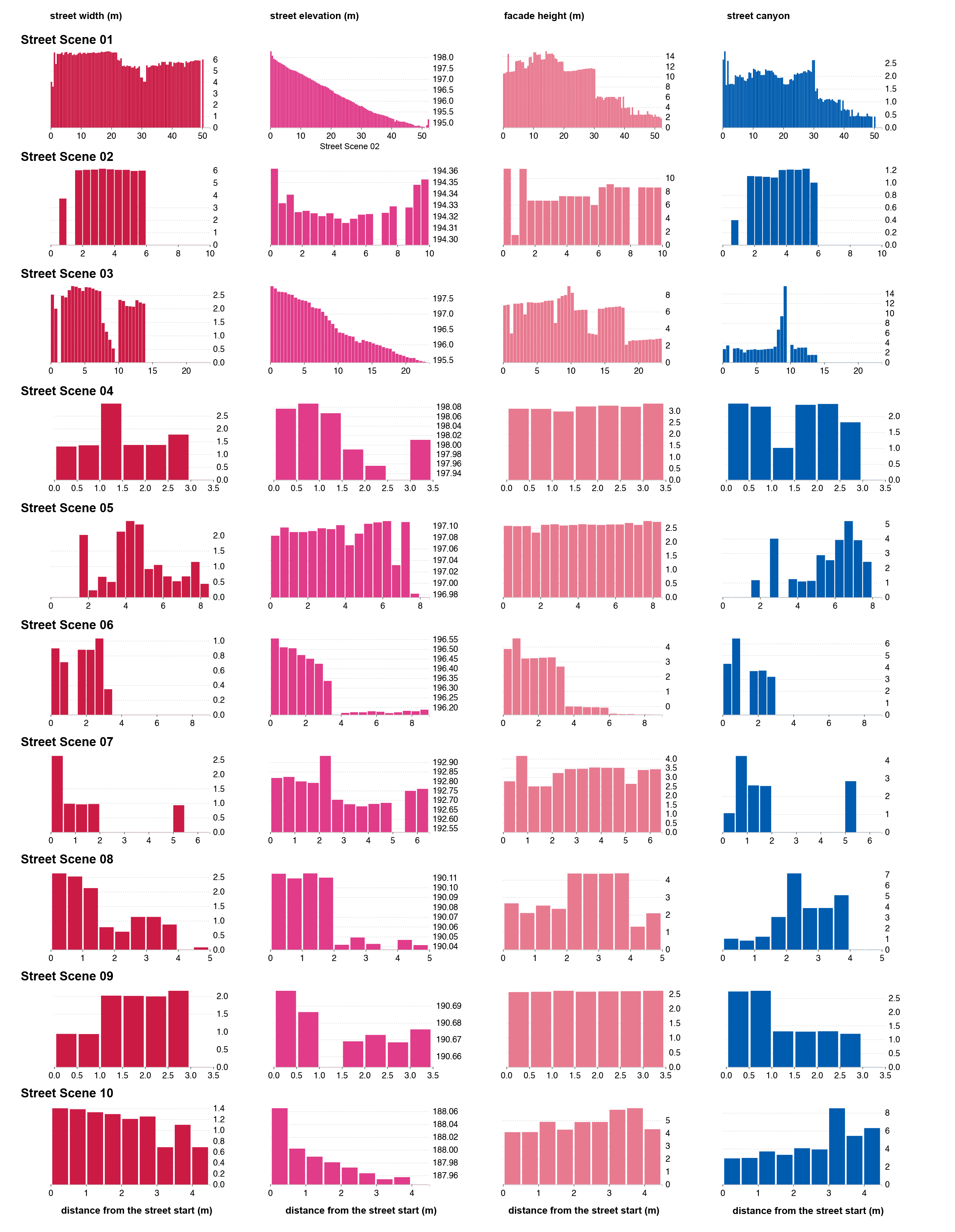}
	\caption{Figure shows the local analysis results for each street scene.}
	\label{local}
\end{figure}

Figure \ref{local} shows the variation in each of the 5 morphological metrics within each street scene. Each bar graph indicates the values for a given metric at half-meter intervals along the primary axis of the street scene. Each row of plots shows the metrics for a given street scene, and each column (and color) of plots represents a different metric, in the following order: street width (red), street elevation (purple), facade height (pink), street canyon (blue). Within each plot, the X axis represents the actual distance in meters from one end of the street to the other, such that the bars along it correspond to the half-meter-wide bins of points into which we divide the street.\footnote{Some plots do not show values for every bin. Several factors can produce these gaps: the Y value for a given bin may be equal to the minimum for the street scene, or there may not be facades present for the bin (calculating street width requires two facades on opposite sides of the street, and calculating facade height requires at least one facade adjacent to the street; street canyon requires both a street width and a facade height for a given bin).} The Y axis of the plot represents the morphological metric value, which we calculate for each half-meter-wide bin.

The figure sheds light on the morphological fluctuations within street scenes: for instance, the plots for street scenes 1 and 3 indicate sloped terrain as elevation steadily declines along the street. Similarly, a comparison of the facade height between street scenes 3 and 6 depicts nuanced changes in the distribution of tall and short buildings along the street. In street scene 3, the pink bars with similar high facade height values represent one building or pair of buildings across the street from one another, and the subsequent segments of smaller bars indicate shorter buildings in the same sense.

The local results also reveal key insights about the spatial configuration of street scenes across the Courtyard and Hillside Scans. Within the Courtyard Scan, for example, the plots show that the street width is very consistent across street scenes 1 and 2. However, despite having similar street width, the street canyon plots show that their street configuration looks very different. In particular, the street canyon values of street scene 2 are much lower than those of street scene 1, suggesting that its a much wider and more open street environment. 

The local results for the Hillside Scan indicate its morphological distinction from the Courtyard Scan. The local results for the Hillside street scenes stay very constant within the street scenes. For example, there is little fluctuation in the street width of street scene 10, or the facade height in street scenes 7 and 9. This might be attributed to the fact that street scenes in this scan tend to be very short (scene 9 is only 3.5 meters long), which leaves very little space for the morphological metrics to vary significantly. 

In summary, the global analysis enables an overall comparison that captures the less-variable profiles of smaller street scenes, while the local analysis allows for an interpretation of the nuanced morphological changes typical of larger and more varied street scenes that may be masked by the global analysis.

\subsection{Mapping Morphology}
\begin{figure}[!ht]
	\centering
	\figuretitle{}
	\includegraphics[width=1\linewidth]{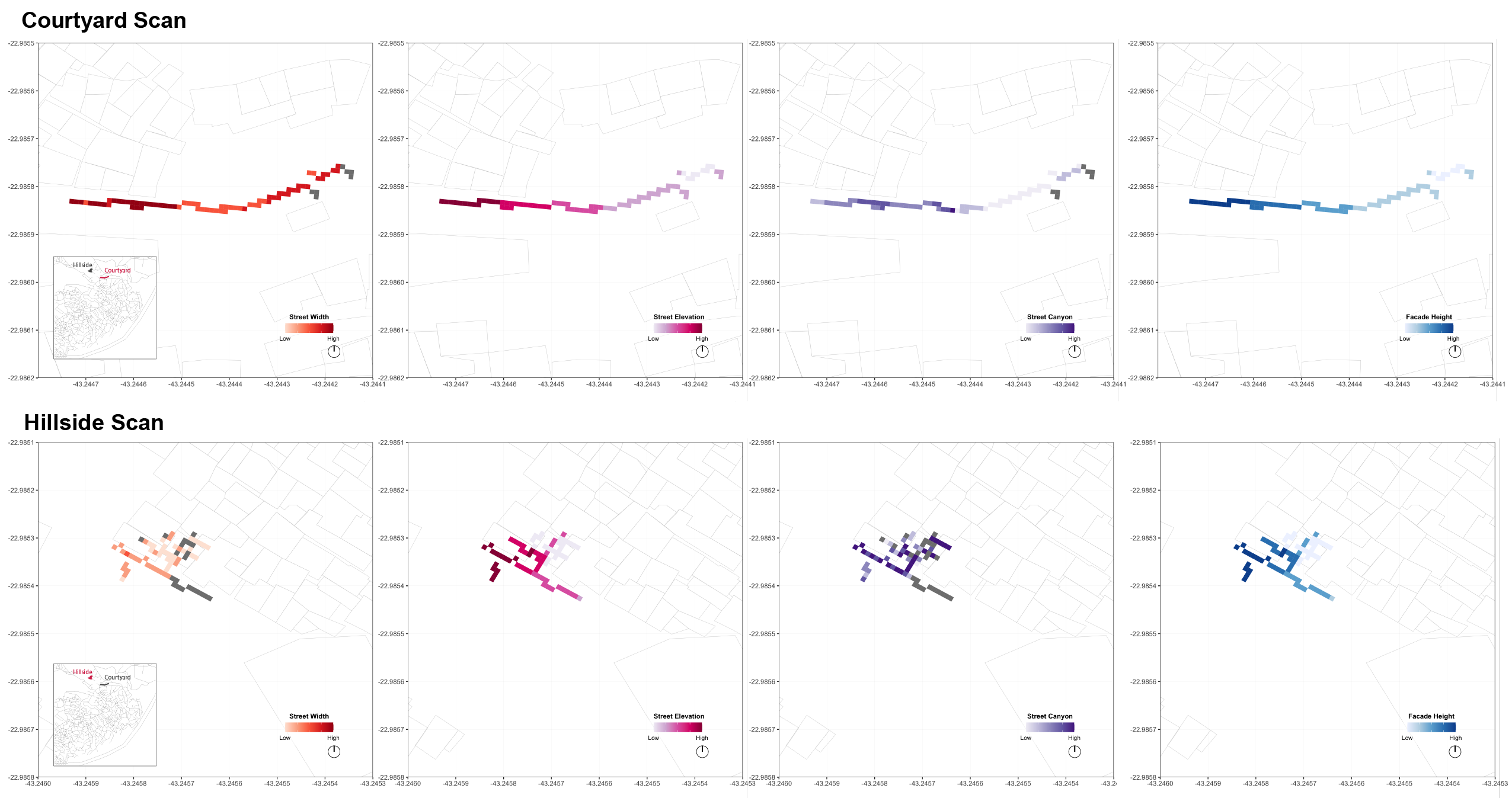}
	\caption{Maps show the local analysis results for each scan, situated within the favela.}
	\label{maps}
\end{figure}

Our local analysis traces morphological variations in the built environment at a high spatial resolution, making it easily translatable from the sphere of abstract analysis to that of concrete urban mapping. To accomplish this transition, we produce maps of each scanning area and superimpose our local analysis measurements as strings of pixels, as shown in Figure \ref{maps}. Each pixel represents one band of points in a street scene. Each pixel is located at the centerline of the street within its band and colored according to the morphological metric value measured for that band, such that darker colors represent higher values. The values used for each metric are normalized to their maximum within each scan. We present a separate map for each morphological metric in each scanning location.

Using these detailed morphological maps, allows us to trace the variation of each metric not only within the extent of a single street scene (as in our local analysis), but also across the larger context of the surrounding built environment. Like our global analysis, these maps are productive for overall comparison of street scenes or areas to one another, but they specify this comparison by separating its components by metric and locating the measurements geographically in space. Thus, this rendering of our results enables simultaneous local and global exploration of morphology in the scanning area.

This final step of mapping our most detailed results onto the favela's plan makes the insights from our analysis clear and actionable. As a set of samples, these maps can show morphological variation across space such as topography, the pinching and widening of streets, and clusters of tall buildings. Extended to the scale of the entire favela, such maps would reveal hotspots of crowding, steep inclines, and dangerous overlaps of the two, enabling city officials and urban planners to pinpoint locations in need of intervention. These maps demonstrate the utility of LiDAR as a means of concisely and legibly articulating informal settlement morphology to the institutions that can act upon it.

\subsection{Measurement Validation}
To validate our morphology measurements we compare our global results with official maps from the Public Works Company of the State of Rio de Janeiro at the locations of the scans, as well as satellite and 3D mesh data from Google Earth. The Public Works maps were collected as part of a participatory planning initiative with the community in 2004 \citep{toledo_2004}.\footnote{These maps are only available for the year 2004. Although they have not been updated recently, they offer the most comprehensive official data source available to ground-truth our data.} We use these plans to confirm our measurements of street width. To ground-truth our elevation measurements, we compared them to a 3D model of Rocinha produced by the Public Works Company of the State of Rio de Janeiro. Recall that the elevation values in the point cloud are established relative to the first point in each scan rather than absolute meters above sea level, therefore we validate elevation in terms of the elevation difference between street scenes rather than as absolute elevation values. Finally, to validate our measurements of facade height and density, we compare them to building counts from satellite images and building heights obtained from Google Earth Pro. Table \ref{validation} in the appendix shows a comparison of our calculations relative to the official sources. Reassuringly, there were not meaningful differences between the two.  

\section{Discussion}\label{discussion}
In this section, we contextualize the findings of our morphological analysis within the present-day urban configuration of the scanning area and the historical development of Rocinha.

Our results align closely with historical evidence tracing the development of the favela. For example, our global analysis indicates that street scenes 1 and 2, which are both in the Courtyard Scan, are much more similar to each other than they are to other street scenes in our sample. The high street width values close to the global maximum can be explained by the formal intervention that widened the courtyard as a social gathering space, and high facade heterogeneity due to the presence of both tall buildings and low retaining walls. Their low facade densities can be explained by the large buildings organized along the spacious courtyard, and their low street canyon values can occur because the scale of the street forces the ratio of facade height to street surface area down. Thus, our morphology metrics capture meaningful properties of the built environment that are consistent with the urban reality of the Courtyard Scan.

Similarly, our results for street scenes in the Hillside Scan capture the spatial dynamics we expect of an informal residential area: compactness, density, and verticality. The facade heterogeneity, facade density, and street canyon values in street scenes 6-10 are markers of a built environment characterized by a variable conglomeration of facades that crowd and tower over narrow streets. These morphological profiles are consistent with the traditions of informal construction in favelas, where unregulated property lines enable buildings to encroach on street space, and the expansion strategy of building additional floors on top of flat-roofed homes results in variable facade heights.

Extending our morphological analysis to the local level enables us to unpack variations and differences that are not immediately apparent in our global analysis. These high-resolution measurements are indicative of sharp changes in the favela environment that occur across short distances: even within a single street scene, the built environment may look very different from one end of the street to the other. This dynamic underscores the importance of high-resolution morphological analysis in informal settlements, and it emphasizes the utility of LiDAR data, whose pointwise and aggregate dimensions enable the interpretation of the favela at multiple levels.

\section{Conclusion}\label{conclusion}
This paper uses terrestrial LiDAR data to analyze the morphology of informal settlements. We take Rocinha, the largest \emph{favela} in Rio de Janeiro, Brazil as a case study to illustrate this application. As a consolidated and well-mapped favela, Rocinha enables us to ground-truth our measurements and thereby validate this approach for replication elsewhere. Our methodology reveals meaningful morphological differences and commonalities both in terms of the global morphology of streets and their local variations. To measure the morphology of the informal settlement we propose a series of five metrics. related to the geometric characteristics of the street and its facades: street width; street elevation; facade heterogeneity; facade density; and street canyon and validate them using official maps and historical evidence of the favela's development. Our approach has two primary advantages. First, from a methodological point of view, the 3D measurements we extracted from LiDAR are flexible in terms of their resolution: the point cloud data can be aggregated and measured at various scales, enabling both global and local analysis. Another advantage of this approach is that it allows us to quantify morphological properties of informal settlements at large, given its automatic and data-efficient nature. Both of these strengths prime our approach for implementation as a survey and management tool for local agencies to determine and execute strategic improvements in the favela. Moreover, morphological analysis of informal settlements provides a basis for the study of important environmental and social issues that impact informal settlements, such as airflow, landslide risk and structural safety, property rights, and accessibility.

Our global and local results demonstrate the benefit of conducting morphological analysis at two different scales and emphasize the utility of the street scene approach. The global results consist of morphological summaries of each street scene, making them best suited to comparing small street scene such as those in the Hillside Scan. Meanwhile, the complexity of larger street scenes simplified in the global analysis is fully expressed in our local analysis, which indicates morphological variation within street scenes. This dual approach to morphological analysis culminates in maps showing the local results for each morphological metric situated in a plan of the favela, which represent our highest resolution measurements in context. All together, these complimentary scales of analysis offer useful comparative insights about morphology for the range of urban conditions in the favela.

This paper provides several fruitful areas for future research that can complement and improve our methodology. First, the LiDAR data we use for the descriptive analysis is only available for two relatively small areas of Rocinha. This means that the insights we can draw about the morphology of the informal settlement are limited to these areas. A full scan of the favela could allow future work to incorporate other complementary morphology metrics, such as connectivity, compactness, and other properties of the settlement that can be best measured at larger scales. Second, our plane extraction model can be improved to better match streets defined by their intersections. Our current approach focuses on extracting large horizontal and vertical planes, and cannot accurately identify very small connected planes such as staircases. This results in a series of small street scenes located at different elevations rather than a continuous street scene connecting them. Access to more LiDAR data and the development of complementary plane extraction procedures could enhance morphological studies looking to characterize the complexity of informal settlements.

\clearpage

{\small \renewcommand{\baselinestretch}{1} \addtolength{\parskip}{.65pt}
               \bibliographystyle{econometrica_etal}
               \bibliography{main.bib}

@article{chokyu2018a,
author = {Chokyu, Margaret Lica and Dias, Maria Angela},
doi = {10.17265/1934-7359/2018.07.002.},
journal = {Journal of Civil Engineering and Architecture},
pages = {483--492},
title = {{Favela's Houses as Design Reference: Using Shape Grammar}},
volume = {12},
year = {2018}
}

@article{barthelemy2014a,
author = {Louf, R{\'{e}}mi and Barthelemy, Marc},
journal = {Physics and Society},
title = {{A typology of street patterns}},
url = {https://arxiv.org/abs/1410.2094.},
year = {2014}
}

@article{Fabricius2008,
author = {Fabricius, Daniela},
journal = {Harvard Design Magazine},
number = {S/S},
title = {{Resisting Representation: The Informal Geographies of Rio de Janeiro}},
volume = {28},
year = {2008}
}

@misc{alexander1977a,
author = {Alexander, Christopher and Ishikawa, Sara and Silverstein, Murray and Jacobson, Max and Fiksdahl-King, Ingrid and Shlomo, Angel},
title = {{A pattern language: Towns, buildings, construction}},
year = {1977}
}

@article{brelsford2018a,
author = {Brelsford, Christa and Martin, Taylor and Hand, Joe and Bettencourt, Lu{\'{i}}s M. A.},
doi = {10.1126/sciadv.aar4644},
journal = {Science Advances},
number = {8},
title = {{Toward cities without slums: Topology and the spatial evolution of neighborhoods}},
volume = {4},
year = {2018}
}

@article{kohli2012a,
author = {Kohli, Divanyi and Sliuzas, Richard and Kerle, Norman and Stein, Alfred},
doi = {10.1016/j.compenvurbsys.2011.11.001},
journal = {Computers, Environment, and Urban Systems},
number = {2},
pages = {154--163},
title = {{An ontology of slums for image-based classification}},
volume = {36},
year = {2012}
}

@article{dovey2020a,
author = {Dovey, Kim and van Oostrum, Matthijs and Chatterjee, Ishita and Shafique, Tanzil},
doi = {10.1016/j.habitatint.2020.102240.},
journal = {Habitat International},
title = {{Towards a morphogenesis of informal settlements}},
volume = {104},
year = {2020}
}

@inproceedings{sobreira2007a,
address = {Brazil},
author = {Sobreira, Fabiano},
booktitle = {Proceedings of the Fourteenth International Seminar on Urban Form, Ouro Preto},
title = {{Favelas, barriadas, bidonvilles: the universal morphology of poverty}},
url = {https://fabianosobreira.files.wordpress.com/2009/07/fsobreira-paper-isuf-2007.pdf.},
year = {2007}
}

@article{jedwab2021a,
author = {Jedwab, Remi and Loungani, Prakash and Yezer, Anthonhy},
doi = {10.1016/j.regsciurbeco.2020.103609},
journal = {Regional Science and Urban Economics},
title = {{Comparing cities in developed and developing countries: Population, land area, building height and crowding}},
volume = {86},
year = {2021}
}

@article{kakon2010a,
author = {Kakon, Anisha Noori and Mishima, Nobuo and Kojima, Shoichi and Yoko, Taguchi},
journal = {American J. of Engineering and Applied Sciences},
number = {3},
pages = {545--551},
title = {{Assessment of Thermal Comfort in Respect to Building Height in a High-Density City in the Tropics}},
url = {https://www.researchgate.net/profile/Shoichi{\_}Kojima/publication/49619577{\_}Assessment{\_}of{\_}Thermal{\_}Comfort{\_}in{\_}Respect{\_}to{\_}Building{\_}Height{\_}in{\_}a{\_}High-Density{\_}City{\_}in{\_}the{\_}Tropics/links/590fca40a6fdccad7b126cae/Assessment-of-Thermal-Comfort-in-Respect-to-Building-},
volume = {3},
year = {2010}
}

@article{gevaert2017a,
author = {Gevaert, Caroline M. and Persello, Claudio and Sliuzas, Richard and Vosselman, George},
doi = {10.1016/j.isprsjprs.2017.01.017.},
journal = {ISPRS Journal of Photogrammetry and Remote Sensing},
pages = {225--236},
title = {{Informal settlement classification using point-cloud and image-based features from UAV data}},
volume = {125},
year = {2017}
}

@incollection{bonczak2019a,
author = {Bonczak, Bartosz and Kontokosta, Constantine E.},
booktitle = {Computers, Environment, and Urban Systems },
doi = {10.1016/j.compenvurbsys.2018.09.004},
pages = {126--142},
title = {{Large-scale parameterization of 3D building morphology in complex urban landscapes using aerial LiDAR and city administrative data}},
volume = {73},
year = {2019}
}

@inproceedings{temba2015a,
author = {Temba, Plinio and Nero, Marcelo A. and Botelho, Lucas M. and Lopes, Matheus E. C.},
booktitle = {Proceedings Volume 9607, Earth Observing Systems XX},
doi = {10.1117/12.2187090.},
title = {{Building vectorization inside a favela using LiDAR spot elevation}},
year = {2015}
}

@article{wang-a,
author = {Wang, Jiong and Kuffer, Monika and Sliuzas, Richard and Kohli, Divyani},
doi = {10.1016/j.scitotenv.2018.09.324.},
journal = {Science of the Total Environment},
number = {2},
pages = {1805--1817},
title = {{The exposure of slums to high temperature: Morphology-based local scale thermal patterns}},
volume = {650},
year = {2019}
}

@misc{guimar2016a,
address = {Rio de Janeiro},
author = {Guitar{\~{a}}es, Saulo Pereira and Taylor, Sheila},
booktitle = {Rio on Watch},
month = {mar},
title = {{Accessibility is a Challenge for Older People in Favelas}},
url = {https://www.rioonwatch.org/?p=27236},
year = {2016}
}

@incollection{ribeiro2019a,
author = {Ribeiro, S and Jarzabeck-Rychard, M and Cintra, J and Maas, H.-G.},
booktitle = {ISPRS Annals of the Photogrammetry, Remote Sensing and Spatial Information Sciences},
doi = {10.5194/isprs-annals-IV-2-W5-437-2019.},
pages = {437--444},
title = {{Describing the vertical structure of informal settlements on the basis of LiDAR data - A case study for favelas (slums) in Sao Paulo City}},
volume = {IV2},
year = {2019}
}

@article{mukeku2018a,
author = {Mukeku, Joseph},
doi = {10.1177/2455747118790581},
journal = {Urbanisation},
number = {1},
pages = {17--32},
title = {{Urban Slum Morphology and Socio-economic Analogies: A Case Study of Kibera Slum, Nairobi, Kenya}},
volume = {3},
year = {2018}
}

@inproceedings{loureiro2017a,
author = {Loureiro, V{\^{a}}nia Raquel Teles and de Medeiros, Val{\'{e}}rio Augusto Soares and Guerreiro, Maria Ros{\'{a}}lia},
booktitle = {Proceedings of the 11th Space Syntax Symposium},
title = {{Configuration of self-organizing informality: socio-spatial dynamic in favelas}},
url = {https://repositorio.iscte-iul.pt/bitstream/10071/14581/1/86.pdf.},
year = {2017}
}

@book{conzen_thinking_2004,
address = {Oxford ; New York},
author = {Conzen, M R G and Conzen, Michael P},
isbn = {978-3-03910-276-1 978-0-8204-7203-4},
publisher = {Peter Lang},
shorttitle = {Thinking about urban form},
title = {{Thinking about urban form: papers on urban morphology, 1932-1998}},
year = {2004}
}

@article{bardhan2018a,
author = {Bardhan, Ronita and Debnath, Ramit and Malik, Jeetika and Sarkhar, Ahana},
doi = {10.1016/j.scs.2018.04.038.},
journal = {Sustainable Cities and Society},
pages = {126--138},
title = {{Low-income housing layouts under socio-architectural complexities: A parametric study for sustainable slum rehabilitation}},
volume = {41},
year = {2018}
}

@misc{loureiro2019a,
author = {Loureiro, V{\^{a}}nia Raquel Teles and de Medeiros, Val{\'{e}}rio Augusto Solares and Guerreiro, Maria Ros{\'{a}}lia da Palma},
booktitle = {ISUF 2019 - International Seminar On Urban Form},
title = {{Sociospatial reading of favela: a comparative analysis from organic Portuguese Cities}},
url = {https://repositorio.unb.br/bitstream/10482/36926/1/EVENTO{\_}SociospatialReadingFavela.pdf.},
volume = {26},
year = {2019}
}

@article{tooke2014a,
author = {Tooke, Thoreau Rory and Coops, Nicholas C. and Webster, Jessica},
doi = {10.1016/j.enbuild.2013.10.004},
journal = {Energy and Buildings},
number = {A},
pages = {603--610},
title = {{Predicting building ages from LiDAR data with random forests for building energy modeling}},
volume = {68},
year = {2014}
}

@article{stark2020a,
author = {Stark, Thomas and Wurm, Michael and Zhu, Xiao Xiang and Taubenb{\"{o}}ck, Hannes},
doi = {10.1109/JSTARS.2020.3018862.},
journal = {IEEE Journal of Selected Topics in Applied Earth Observations and Remote Sensing},
pages = {5251--5263},
title = {{Satellite-Based Mapping of Urban Poverty With Transfer-Learned Slum Morphologies}},
volume = {13},
year = {2020}
}

@article{duque2017a,
author = {Duque, Juan C. and Patino, Jorge E. and Betancourt, Alejandro},
doi = {10.3390/rs9090895.},
journal = {Remote Sensing},
number = {9},
title = {{Exploring the Potential of Machine Learning for Automatic Slum Identification from VHR Imagery}},
volume = {9},
year = {2017}
}

@article{taubenboeck2018a,
author = {Taubenb{\"{o}}ck, Hannes and Kraff, Nicholas Johannes and Wurm, Michael},
doi = {10.1016/j.apgeog.2018.02.002.},
journal = {Applied Geography},
pages = {150--167},
title = {{The morphology of the Arrival City - A global categorization based on literature surveys and remotely sensed data}},
volume = {92},
year = {2018}
}

@article{zappulla2014a,
author = {Zappulla, Carmelo and Suau, Cristiao and Fikfak, Alenka},
doi = {10.3846/20297955.2014.987368.},
journal = {Journal of Architecture and Urbanism},
number = {4},
pages = {247--264},
title = {{The pattern making of mega-slums on semantics in slum urban cultures}},
volume = {38},
year = {2014}
}

@misc{sliuzas2017a,
address = {Dubai},
author = {Sliuzas, Richard and Kuffer, Monika and Gevaert, Caroline and Persello, Claudio and Pfeffer, Karin},
booktitle = {Joint Urban Remote Sensing Event (JURSE)},
doi = {10.1109/JURSE.2017.7924589.},
pages = {1--4,},
title = {{Slum mapping}},
year = {2017}
}

@article{verniz2020a,
author = {Verniz, Debora and Duarte, Jos{\'{e}} P.},
doi = {10.1177/2399808319897625.},
journal = {Environment and Planning B: Urban Analytics and City Science},
title = {{Santa Marta Urban Grammar: Unraveling the spontaneous occupation of Brazilian informal settlements}},
year = {2020}
}

@article{schnabel2007a,
address = {Oxford, UK},
author = {Schnabel, Ruwen and Wahl, Roland and Klein, Reinhard},
doi = {10.1111/j.1467-8659.2007.01016.x},
journal = {Computer graphics forum},
number = {2},
pages = {214--226},
publisher = {Blackwell Publishing Ltd},
title = {{Efficient RANSAC for point-cloud shape detection}},
volume = {26},
year = {2007}
}

@article{jana2020a,
author = {Jana, Arnab and Sarkar, Ahana and Bardhan, Ronita},
doi = {10.1016/j.landusepol.2020.105052},
journal = {Land Use Policy},
title = {{Analysing outdoor airflow and pollution as a parameter to assess the compatibility of mass-scale low-cost residential development}},
volume = {99},
year = {2020}
}

@article{fischler1981a,
author = {Fischler, Martin A. and Bolles, Robert C.},
journal = {Communications of the ACM},
number = {6},
pages = {381--395},
title = {{Random sample consensus: a paradigm for model fitting with applications to image analysis and automated cartography}},
volume = {24},
year = {1981}
}

@article{dovey2013a,
author = {Dovey, Kim},
doi = {10.1002/ad.1679.},
journal = {Architectural Design},
number = {6},
title = {{Informalising Architecture: The Challenge of Informal Settlements}},
volume = {83},
year = {2013}
}

@article{sandholz2018a,
author = {Sandholz, Simone and Lange, Wolfram and Nehren, Udo},
doi = {10.1016/j.ijdrr.2018.01.020},
journal = {International Journal of Disaster Risk Reduction},
pages = {75--86},
title = {{Governing green change: Ecosystem-based measures for reducing landslide risk in Rio de Janeiro}},
volume = {32},
year = {2018}
}

@incollection{kuffer2014a,
author = {Kuffer, Monika and Barros, Joana and Sliuzas, Richard V.},
booktitle = {Computers, Environment, and Urban Systems},
doi = {10.1016/j.compenvurbsys.2014.07.012},
pages = {138--152},
title = {{The development of a morphological unplanned settlement index using very-high-resolution (VHR) imagery}},
volume = {48},
year = {2014}
}

@article{stiny2015a,
author = {Knight, Terry and Stiny, George},
doi = {10.1016/j.destud.2015.08.006.},
journal = {Design Studies},
number = {A},
pages = {8--28},
title = {{Making grammars: From computing with shapes to computing with things}},
volume = {41},
year = {2015}
}

@article{boeing2021a,
author = {Boeing, Geoff},
doi = {10.1016/j.ijinfomgt.2019.09.009.},
journal = {International Journal of Information Management},
title = {{Spatial information and the legibility of urban form: Big data in urban morphology}},
volume = {56},
year = {2021}
}

@incollection{kamalipour2020a,
author = {Kamalipour, Hesam and Dovey, Kim},
booktitle = {Habitat International},
doi = {10.1016/j.habitatint.2020.102133.},
title = {{Incremental production of urban space: a typology of informal design}},
volume = {98},
year = {2020}
}

@article{wurm2019a,
author = {Wurm, Michael and Stark, Thomas and Zhu, Xiao Xiang and Weigand, Matthias and Taubenb{\"{o}}ck, Hannes},
doi = {10.1016/j.isprsjprs.2019.02.006.},
journal = {ISPRS Journal of Photogrammetry and Remote Sensing},
pages = {59--69},
title = {{Semantic segmentation of slums in satellite images using transfer learning on fully convolutional neural networks}},
volume = {150},
year = {2019}
}

@article{prabhu2021a,
author = {Prabhu, R. and Parvathavarthini, B. and Raja, R. A.},
doi = {10.1080/01431161.2020.1834167.},
journal = {International Journal of Remote Sensing},
number = {1},
pages = {172--190},
title = {{Slum Extraction from High Resolution Satellite Data using Mathematical Morphology based approach}},
volume = {42},
year = {2020}
}

@article{cavalcanti2017a,
author = {Cavalcanti, Ana Rosa Chagas},
journal = {Traditional Dwellings and Settlements Review},
number = {2},
pages = {71--81},
title = {{Work, Slums, and Informal Settlement Traditions: Architecture of the Favela Do Telegrafo}},
url = {http://www.jstor.org/stable/44779812.},
volume = {28},
year = {2017}
}

@article{shan2019a,
author = {Shan, Li and Petrone, Floriana and Kunnath, Sashi},
doi = {10.1016/j.engstruct.2019.01.052.},
journal = {Engineering Structures},
pages = {690--701},
title = {{Robustness of RC buildings to progressive collapse: Influence of building height}},
volume = {183},
year = {2019}
}

@article{scott2017a,
author = {Scott, Anna A. and Misani, Herbert and Okoth, Jerrim and Jordan, Asha and Gohlke, Julia and Ouma, Gilbert and Arrighi, Julie and Zaitchik, Ben F. and Jjemba, Eddie and Verjee, Safia and Waugh, Derren W.},
doi = {10.1371/journal.pone.0187300.},
journal = {PLoS ONE},
number = {11},
title = {{Temperature and heat in informal settlements in Nairobi}},
volume = {12},
year = {2017}
}

@incollection{a2019a,
author = {Martins, A. Nuno and Farias, Jacira Saavedra},
booktitle = {Sustainable Development},
doi = {10.1002/sd.1879},
pages = {205--213},
title = {{Inclusive sustainability within favela upgrading and incremental housing: The case of Rocinha in Rio de Janeiro}},
volume = {27},
year = {2019}
}

@article{schmitt2018a,
author = {Schmitt, Andreas and Sieg, Tobias and Wurm, Michael and Taubenb{\"{o}}ck, Hannes},
doi = {10.1016/j.jag.2017.09.006.},
journal = {International Journal of Applied Earth Observation and Geoinformation},
pages = {181--198},
title = {{Investigation on the separability of slums by multi-aspect TerraSAR-X dual-co-polarized high resolution spotlight images based on the multi-scale evaluation of local distributions}},
volume = {64},
year = {2018}
}

@book{moudon_built_1989,
address = {Cambridge, Mass.},
author = {Moudon, Anne Vernez},
isbn = {978-0-262-63120-4},
publisher = {MIT Press},
shorttitle = {Built for change},
title = {{Built for change: neighborhood architecture in San Francisco}},
year = {1989}
}

@article{siksna_effects_1997,
author = {Siksna, Arnis},
issn = {1027-4278},
journal = {Urban Morphology},
pages = {19--33},
title = {{The effects of block size and form in North American and Australian city centres}},
volume = {1},
year = {1997}
}

@misc{toledo_2004, 
place={Rio de Janeiro},
title={Plano Diretor Socio-Espacial da Rocinha Inovando Com Velhas Ideias}, 
publisher={Secretaria Estadual de Obras, Empresa de Obras Publicas }, 
author={Toledo, Luiz Carlos}, 
year={2004}
}
}

\clearpage
\section*{Appendix}
\renewcommand{\thefigure}{A\arabic{figure}}
\setcounter{figure}{0}
\renewcommand{\thetable}{A\arabic{table}}
\setcounter{table}{0}

\textbf{RANSAC Shape detection procedure:}
We use the RANSAC shape detection method, introduced by Schnabel et al (2007) to extract shape models from unorganized point clouds. The algorithm to extract planes works as follows:

Given a point cloud $\mathcal{P} = \{p_1,\dots,p_N\}$ with associated normals $\{n_1,\dots,n_N\}$, the output of the algorithm are planes $\{\Psi_1,\dots,\Psi_n\}$ with its corresponding set of disjoint points $\{\mathcal{P}_{\Psi_1},\dots,\mathcal{P}_{\Psi_n}\}$. To construct the planes, the algorithm first extracts a set of points $\mathcal{P}_{\Psi}$ that belong to a plane $\Psi$ from a all the points remaining in the point cloud $\mathcal{R}$ ($\mathcal{R} \subset \mathcal{P}$). To do this, it calculates the parameters of a candidate plane $\Psi_m$ from $3$ randomly selected points $\{p_1,p_2,p_3\}$ (every combination of 3 points in $\mathcal{R}$ constitute a minimal set that can define a candidate plane), and then finds the corresponding set of points $\mathcal{P}_{\Psi_m}$ that lie within a given distance $\epsilon$ to this candidate plane ($\epsilon$ is a ratio of the point cloud bounding box width, we use a value of 0.0045\textemdash the typical value of this parameter for a single building is 0.002, but since informal settlements are built piecewise, allow a ratio of 0.0045 to allow more variance on the output planes). We run this random selection process for $T$ iterations (the iterations $T$ are required to find a shape with minimum support points $\tau$ in subset $\mathcal{R}$ with an overlooking probability $p_t$. We set $p_\tau$ to 0.1, which means we have a $90\%$ probability of detecting the best model. Recall that our goal is to extract the relevant points, not to find the best model for each plane, so we limit the number of iterations. The best candidate plane constitute the largest connected component on the shape and is chosen using the number of inlier points, the points in $\mathcal{P}_{\Psi_{m}}$ that do not deviate from the plane normal for more than the maximum deviation angle $\alpha$ (we use a normal deviation angle $\alpha$ of 12.5 - 35.0 ($\epsilon$ / 0.012) to adjust for the fact that buildings vary widely in scale in our context \textemdash a typical value $\alpha$ = 20 is used model focused on a single building). The best plane candidate $\Psi_{m_b}$ is only accepted when the number of points $|\mathcal{P}_{\Psi_{m_b}}|$ is no less than the threshold $\tau$(we set $\tau$ to 750 in order to include small planes from buildings, but exclude urban furniture). Once a best plane model is found, the corresponding set of points $\mathcal{P}_{\Psi_{m_b}}$ are removed from the point cloud $\mathcal{P}$ and the algorithm is repeated until the only remaining points are noise that do not constitute a plane with minimum support points. In addition, the algorithm also requires a sampling resolution $r$, which defines how the algorithm samples points from the point cloud, we set $r$ to be 0.05m, corresponding to the distance between neighboring points in the data. We choose this number to balance the risk of classifying more noise planes but blending groups of small planes together as one large plane (larger values) and classifying fewer noise planes but having large streets or facades divided into small planes (smaller values). A summary of the parameters we used is shown in Table A.

\begin{table}[!t]
	\centering
	\caption\\\vspace{0.3cm}{\textsc{\small{Parameters used in the RANSAC Procedure}}}\vspace{0.2cm}
	\label{ransac_table1}
	\resizebox{\linewidth}{!}{\begin{tabular}{L{3cm}L{4cm}L{4cm}C{3cm}C{3cm}C{3cm}}\toprule\toprule
			\multicolumn{1}{c}{Parameters} &\multicolumn{1}{c}{Definition} & \multicolumn{1}{c}{Reasoning} & \multicolumn{1}{c}{Hillside Scan}  & \multicolumn{1}{c}{Courtyard Scan} \\
			\cmidrule(r){1-1}\cmidrule(r){2-2}\cmidrule(r){3-3}\cmidrule(r){4-4}\cmidrule(r){5-5}
			Minimum Support Points $\tau$ & Minimum number of points required for a point cloud fragment to be considered a plane candidate & We use a value that allows us to include building facades but exclude urban furniture & 750 & 750 \\
			Maximum distance to primitive $\epsilon$ & Maximum distance of sample points to plane to be included in the corresponding set of points of the plane, given as a ratio of bounding box width & The value has a linear relationship with maximum bounding box width, so that more variance is allowed for larger areas & 0.15 & 0.3 \\
			Sampling resolution $r$ (m) & Distance between neighbour sampling points, equivalent to the point cloud resolution & the value we use corresponds to the scan resolution & 0.05 & 0.05  \\
			Maximum normal deviation $\alpha$ (degrees) & Maximum deviation of the normal of initial randomly selected points from the normal of the plane defined by them & Calculated based on a linear relationship with maximum distance to primitive & 12.5 & 25.0 \\
			Overlooking Probability $p_t$ & The probability that no other better primitive candidates are overlooked (the chosen candidate is the best) during sampling. & We selected a probability value that balances between the overall iterations and best plane model & 0.1 & 0.1  \\
			\\\bottomrule
	\end{tabular}}
	\begin{minipage}{\linewidth}
		\scriptsize Notes: The Table shows the parameters used for the RANSAC calibration. Column 1 shows the parameter used. Column 2 and 3 describes the parameter and provide a brief intuition of its use. Columns 4 and 5 show the parameter values used for the hillside and the courtyard scan, respectively. 
	\end{minipage}
\end{table}	

\begin{figure}[!t]
	\centering
	\figuretitle{}
	\includegraphics[width=1\linewidth]{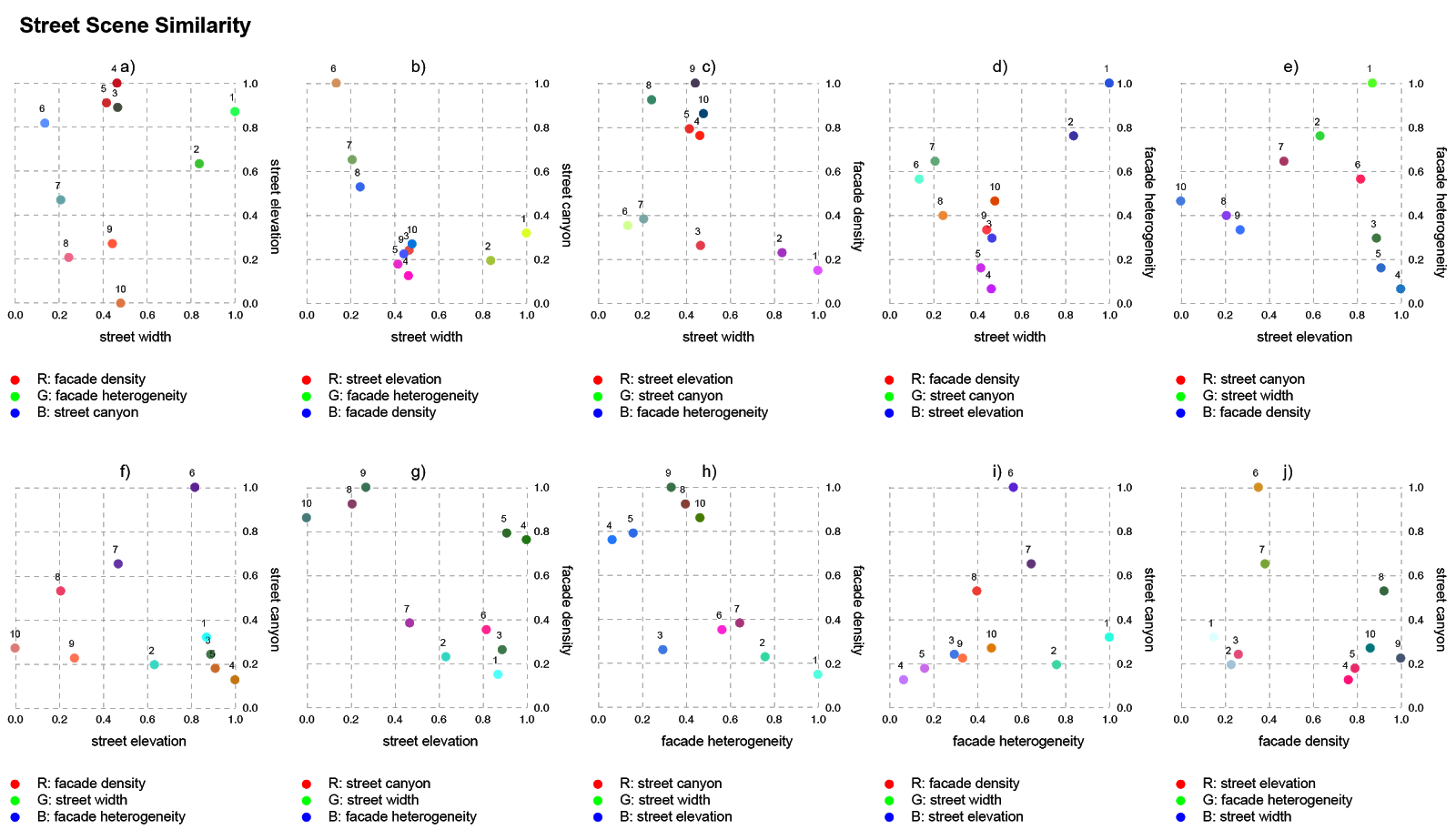}
	\caption{Figure shows global analysis as comparative plots in 5 dimensions (X, Y axes, and R, G, B colors), where each dimension is a morphological metric.}
	\label{comparative}
\end{figure} 

\begin{table}[!t]
	\centering
	\caption\\\vspace{0.3cm}{\textsc{\small{Ground Truth Metric Values}}}\vspace{0.2cm}
	\label{validation}
	\resizebox{\linewidth}{!}{\begin{tabular}{L{3cm}L{4cm}L{4cm}C{3cm}C{3cm}C{3cm}}\toprule\toprule
			\multicolumn{1}{c}{Street Scene Number} &\multicolumn{1}{c}{Street Width} & \multicolumn{1}{c}{Elevation} & \multicolumn{1}{c}{Maximum Building Height}  & \multicolumn{1}{c}{Building Count} \\
			\cmidrule(r){1-1}\cmidrule(r){2-2}\cmidrule(r){3-3}\cmidrule(r){4-4}\cmidrule(r){5-5}
			Source & Rocinha Masterplan (Public Works Company of the State of Rio de Janeiro) & Rocinha 3D plan (Public Works Company of the State of Rio de Janeiro) & Google Maps Satellite & Google Earth Pro 3D Mesh \\
			1 & 6.21 m & 177 m & 18.23 m & 12 \\
			2 & 5.78 m & 172 m & 8.71 m & 3 \\
			3 & 3.29 m & 193 m & 10.28 m & 6 \\
			4 & 2.67 m & 198 m & 2.62 m & 2 \\
			5 & 3.41 m & 187 m & 3.61 m & 2 \\
			6 & 0.81 m & 189 m & 9.42 m & 2 \\
			7 & 1.94 m & 190 m & 6.03 m & 2  \\
			8 & 1.73 m & 181 m & 6.35 m & 2 \\
			9 & 2.29 m & 182 m & 5.21 m & 2 \\
			10 & 2.10 m & 177 m & 3.86 m & 3 \\
			\\\bottomrule
	\end{tabular}}
	\begin{minipage}{\linewidth}
		\scriptsize Notes: The Table shows the ground truth information used to validate the street scene measurements. Each column refers to a different ground truth metric. Row 1 shows the source used to gather the values for each metrics, and rows 2-11 show the values identified for each street scene. This information confirms the validity of the street scene analysis, which does not differ significantly from the ground truth. 
	\end{minipage}
\end{table}	

\end{document}